\documentclass[letterpaper, 10 pt, conference]{ieeeconf}  

\IEEEoverridecommandlockouts                              
\usepackage{algorithmicx}
\usepackage{multicol}
\usepackage[bookmarks=true]{hyperref}

\usepackage{times}
\usepackage{epsfig}
\usepackage{graphicx}
\usepackage{amsmath}
\usepackage{amssymb}
\usepackage{url}
\usepackage{pst-all}
\usepackage{bm}
\usepackage{algorithm}
\usepackage[noend]{algpseudocode}
\usepackage{bbm}
\usepackage{enumerate}
\usepackage{subcaption}
\usepackage{multirow}
\usepackage{hhline}
\usepackage{graphics}
\usepackage{pifont}
\usepackage{color}
\usepackage{booktabs}
\usepackage{gensymb}

\newcommand{\etal}{\mbox{\emph{et al.\ }}}

\begin{document}

\title{Non-parametric Memory for Spatio-Temporal Segmentation\\of Construction Zones for Self-Driving}
\author{
  Min Bai$^{1,2}$ \quad Shenlong Wang$^{1,2}$\\
  Kelvin Wong$^{1,2}$ \quad Ersin Yumer$^{1}$ \quad Raquel Urtasun$^{1,2}$\\
 $^{1}$Uber Advanced Technologies Group \quad $^{2}$University of Toronto\\
 \small\texttt{\{mbai3,slwang,kelvin.wong,yumer,urtasun\}@uber.com}
}

\newcommand{\bx}{\mathbf{x}}
\newcommand{\bb}{\mathbf{b}}
\newcommand{\ba}{\mathbf{a}}
\newcommand{\bo}{\mathbf{o}}
\newcommand{\bz}{\mathbf{z}}
\newcommand{\bp}{\mathbf{p}}
\newcommand{\bn}{\mathbf{n}}
\newcommand{\bw}{\mathbf{w}}
\newcommand{\cI}{\mathcal{I}}
\newcommand{\cF}{\mathcal{F}}
\newcommand{\cL}{\mathcal{L}}
\newcommand{\cG}{\mathcal{G}}
\newcommand{\cM}{\mathcal{M}}
\newcommand{\cS}{\mathcal{S}}
\newcommand{\cT}{\mathcal{T}}
\newcommand{\by}{\mathbf{y}}
\newcommand{\ut}{^{(t)}}
\newcommand{\up}{^{(t-1)}}
\newcommand{\bt}{\mathbf{t}}

\newcommand{\mlane}{\mathrm{lane}}
\newcommand{\mgps}{\mathrm{gps}}
\newcommand{\msign}{\mathrm{sign}}

\newcommand{\dyn}{\textsc{Dyn}}
\newcommand{\online}{\textsc{o}}
\newcommand{\map}{\textsc{m}}
\newcommand{\mask}{\textsc{mask}}

\newcommand{\ersin}[1]{\textcolor{blue}{Ersin: #1}}
\newcommand{\raquel}[1]{\textcolor{red}{Raquel: #1}}
\newcommand{\shenlong}[1]{\textcolor{magenta}{ #1}}
\newcommand{\todo}[1]{\textcolor{red}{To-Do: #1}}
\newcommand{\new}[1]{\textcolor{red}{#1}}

\twocolumn[{%
\renewcommand\twocolumn[1][]{#1}%
\maketitle
\begin{center}
    \centering
    \vspace{-0.7cm}
    \includegraphics[width=0.99\linewidth]{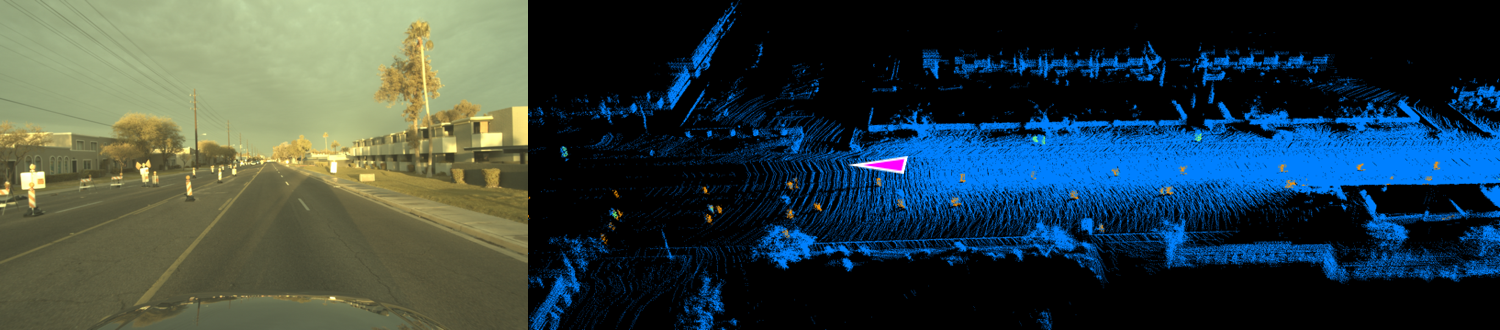}
   \vspace{-0.1cm}
    \captionof{figure}{Left: The current view of from the front camera of an autonomous vehicle (AV). Right: Corresponding memory in our system (Purple arrow shows the direction of travel). Note the denser representation behind the AV, where observations from multiple LiDAR scans and camera images from previous time steps have been contributing. }
    \label{fig:teaser}
    \vspace{-0.2cm}
\end{center}%
}]



\begin{abstract}
In this paper, we introduce a non-parametric memory representation for spatio-temporal segmentation that captures the local space and time around an autonomous vehicle (AV). 
Our representation has three important properties: (i) it {\it remembers} what it has seen in the past, (ii) it {\it reinforces} and  (iii) {\it forgets} its past beliefs based on new evidence. 
Reinforcing is important as the first time we see an element we might be uncertain, e.g, if the element is heavily occluded or at range.
Forgetting is desirable, as otherwise false positives will make the self driving vehicle behave erratically. 
Our  process is informed by 3D reasoning, as occlusion is key to distinguishing between the desire to forget and to  remember. 
We show how our method can be used as an online component to complement static world representations such as HD maps by detecting and remembering changes that should be superimposed on top of this static view due to such events. 

\end{abstract}


\section{Introduction}

Current approaches to autonomous vehicles (AVs) exploit prior knowledge about the static world by building very detailed maps (HD maps) that include not only roads, buildings, bridges and landmarks, but also traffic lanes, signs, and lights to centimeter accurate 3D representations.
This is used at runtime for localization, traffic light and signage detection, motion and behavior forecasting, as well motion planning. 

Most AV systems assume that the HD map is accurate and up to date. However, this is often not true in the real world due to events like construction, temporary lane closures, lane repainting, and speed limit changes, and so on. 
Thus, online reasoning about map changes due to construction elements and temporary signs is critical for safe operation. This is a difficult problem, as these objects are typically small and may be occluded at the current time. 
Additionally, the vehicle must reliably observe the elements at long distances as seen in Figure \ref{fig:teaser}, where the elements may appear as only a few pixels in the camera image and very few 3D points in the LiDAR measurement. 
Any small misalignment in calibration can also result in impactful offsets at range. This is in stark contrast with many methods designed for indoor environments, where the RGB-D sensors provide very dense and relatively low noise information. 

\begin{figure}
    \begin{center}
        \includegraphics[width=0.99\linewidth]{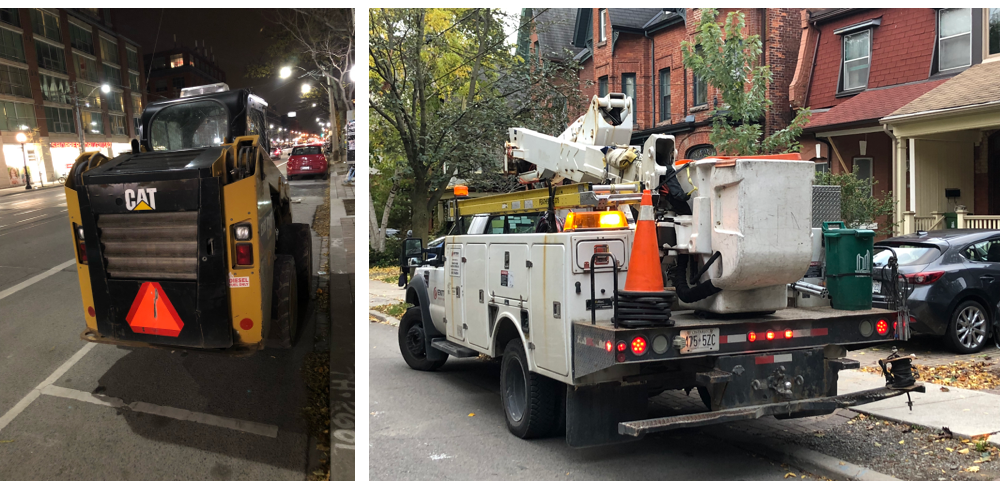}
    \end{center}
    \vspace{-0.3cm}
    \caption{Typical examples of construction sign (left) and construction cone (right) that should be forgotten. As the vehicle to which they are attached may move, past 3D detections of these elements should be forgotten.}
    \label{fig:motivation}
    \vspace{-0.7cm}
\end{figure}

For typical construction elements and road signs, early observations of them may be uncertain, as the element may be heavily occluded or at long range, resulting in even less reliable measurements. We drive our intuition to approach this problem from the following three insights:
(1) As we approach the object, additional confirming information can be used to reinforce the system's belief of an object's presence. 
(2) On the other hand, if new information contradicts previous observations, the system must forget these false positives. Figure \ref{fig:motivation} shows typical examples where construction signs/objects should be forgotten, as they are attached to vehicles whose current positions may differ from where they were previously observed. 
(3) Finally, if there is an obstacle occluding our view of a previously observed element, the lack of new observation should not influence the earlier conjectures. In this case, we should remember the previous state. 
Therefore, the system must correctly aggregate information from multiple time instances to improve robustness of detection. 

Based on these insights, we introduce a non-parametric memory representation that captures the local world around the AV. 
Our proposed model exploits the camera to segment the classes of interest (e.g., construction elements, signs) and the LiDAR to localize the elements in 3D space and for occlusion reasoning. 
At each time step, we use a 2D segmentation model that is trained to segment construction objects and traffic signs in the camera image. 
This observation is associated with the 3D point cloud provided by the LiDAR sensor to localize the object in 3D for the current time step. 
Then, we utilize the non-parametric memory to store the up-to-date semantic predictions of the world around the AV. 
This representation is amenable to both semantic fusion, and spatio-temporal fusion which lets us propagate beliefs both from the semantic understanding in the camera space into the 3D space, as well as from previous time-steps spatio-temporally. 
Compared to semantic mapping methods~\cite{wolf2008semantic, kundu2014joint, schonberger2018semantic} that use a pre-defined update rule to fuse online perceived information with past estimates, we use a learned model to aggregate information that takes into account factors that influence noise and occlusion. 
This allows us to handle cases with mis-alignment, noisy perceived semantics, distance small objects and dynamic changes in a smarter manner.
  
The memory is a dynamically sized graph containing the likely foreground 3D points and their initial classification probabilities from the image segmentation model from the past and present. 
The size of the graph changes with new LiDAR sweeps, as well as deprecation of past information as appropriate. 
Furthermore, we compute an occlusion feature based on the present sweep to determine whether objects observed in the past are now occluded. 
Finally, we combine the above information using a continuous convolution network to segment the available LiDAR points.

We validate our approach on a large dataset captured while driving through construction zones in several cities across North America, consisting of over 4000 snippets with different construction elements such as cones and signs. 
We show that our approach outperforms the baselines by a significant margin.


\section{Related Work}

\paragraph{Image Segmentation}

Recent semantic segmentation approaches successfully exploit deep convolutional neural nets ~\cite{zhao2017pyramid,long2015fully}.  

End-to-end trainable structured prediction techniques model dependencies between the predictions at multiple pixels ~\cite{chen2014semantic,zheng2015conditional,arnab2016higher}. We draw inspiration from this to refine our segmentation results by reasoning about spatio-temporally related 3D points. 
\paragraph{Point Cloud Segmentation}
Initial attempts for 3D point cloud segmentation ~\cite{klasing2008clustering,douillard2011segmentation} used clustering methods to identify individual objects. 
Convolutional networks using a layered 2D representation for the point cloud in bird's-eye view have also been proposed~\cite{li2016vehicle}, where the sparsity of the data can be further exploited ~\cite{ren2018sbnet}. Additionally, voxelization of point clouds is popular. Reductions to computational cost by casting voxel segmentation as 2D convolutions~\cite{chris} or improved memory efficiency ~\cite{octnet} have been proposed. 
Some recent works directly learn convolutional networks over raw point clouds for segmentation. 
The pioneering work~\cite{pointnet} aggregates global and point-wise deep features to predict class-wise labels. It was later extended to a hierarchical aggregation framework~\cite{pointnetplus}. 
An alternative approach is to convert point coordinates to an intermediate space in which convolution operators are possible such as permutohedral lattice~\cite{splatnet}, tangent space~\cite{tangentconv} or over graphs~\cite{3dgnn}. ~\cite{huang2018recurrent} aggregates information across different spatial axis of a single observation using a recurrent structure. 
Recently, parametric continuous convolutions~\cite{wang2018deep} were shown to significantly improve over the state-of-the-art by reasoning directly over neighborhood graphs constructed from points with their natural coordinates in continuous domain. 
However, these techniques use single observations, while our proposed method uses continuous convolutions and occlusion reasoning to fuse information in the memory with current observations to directly merge accumulated past beliefs with new observations at each time step. 

\paragraph{Temporal Reasoning in Segmentation}
Due to the sparse nature of LiDAR point clouds, temporal reasoning by accumulating multiple observations over multiple timesteps have shown to be more robust~\cite{petrovskaya2009model,azim2012detection}. Exploiting all three cues (spatial, semantic, and temporal) within a Bayesian framework, Held~\etal~\cite{held2016probabilistic} demonstrated higher performance gains. Tokmakov~\etal~\cite{tokmakov2018learning} utilized a recurrent convolutional unit to realize a visual memory in image space, where the end result is used for segmenting objects in a video stream. 

\paragraph{Sensor Fusion}
The majority of the earlier work in sensor fusion focused on early fusion where depth data is used as an additional channel in the 2D image space~\cite{silberman2012indoor,gupta2015indoor,eigen2015predicting}. While early fusion appears promising in indoor scenes with high resolution depth sensors, LiDAR point clouds are very sparse, reducing the efficacy of early fusion frameworks. 
More recent end-to-end approaches exploit fusion at multiple levels downstream in the network architecture where the point cloud is rasterized into a birds-eye view ~\cite{chen2017multi,ku2018joint,liang2018deep}. ~\cite{gellert2018deep} fused a bird's-eye view projection of a camera image with the top down view of the point cloud for lane detection. 
Qi~\etal~\cite{qi2017frustum} use a late fusion strategy with a two-stage pipeline where 2D detections in the image space is used as a hard prior in the form of a frustum to further perform binary segmentation of the foreground object in the point cloud. 
Our method leverages both  camera and LiDAR data. However, contrary to the aforementioned approaches, we exploit an external memory which results in robust modelling of static objects that allows us to explicitly remember existing objects rather than re-discovering them with the next observation.

\paragraph{Dynamic Semantic Mapping}
Building dynamically changing semantic maps is a active topic in robotics. It is crucial for the robot to complete many tasks in a dynamic evolving environment, such as navigation \cite{wolf2008semantic}, localization \cite{schonberger2018semantic} and manipulation \cite{galindo2008robot}. There is a large body of work in semantic 3D reconstruction using RGB-D data for indoor scenes~\cite{hermans2014dense,stuckler2014multi,mccormac2017semanticfusion}. Cameras and LiDAR have been also exploited to build semantics maps from urban environment \cite{wolf2008semantic, kundu2014joint, schonberger2018semantic}. Most of these methods use a pre-defined update rule to fuse online perceived information with past estimates. However, the optimal trade-off between these two sources in a dynamic and noisy scene is difficult to encode. On the other hand, we use a learned model to aggregate information that takes into account factors that influence noise and occlusion. This allows us to handle cases with mis-alignment, noisy perceived semantics, distant small objects and dynamic changes in a smarter manner.

\paragraph{Memory Networks}
The majority of approaches utilizing memory rely on recurrent neural networks based on a long short-term memory (LSTM) ~\cite{tokmakov2018learning,stollenga2015parallel}, where the memory is represented as a  1D-vector. More recently, ~\cite{romera2016recurrent,ren2017end} have shown techniques using ConvLSTMs to explicitly remember spatial information. Such convolutional approaches to LSTM encoding exploits the spatial relationships in the image space. However, they do not explicitly encode the 3D structure. 
Explicit external memory with differentiable write and read operations~\cite{graves2016hybrid} have been exploited as an external data structure in settings where a direct mapping between the memory and physical quantities is missing. 
In contrast, our memory representation is an explicitly external 4D spatio-temporal buffer that both has a direct mapping to the real world in a local coordinate frame, as well as being compatible with the continuous convolutions~\cite{wang2018deep} for sparse computation in that space.  Our memory is essentially a graph of "points", which is a different memory representation than existing approaches which rely more on canvases. 


\begin{figure*}
    \begin{center}
        \includegraphics[width=0.97\linewidth]{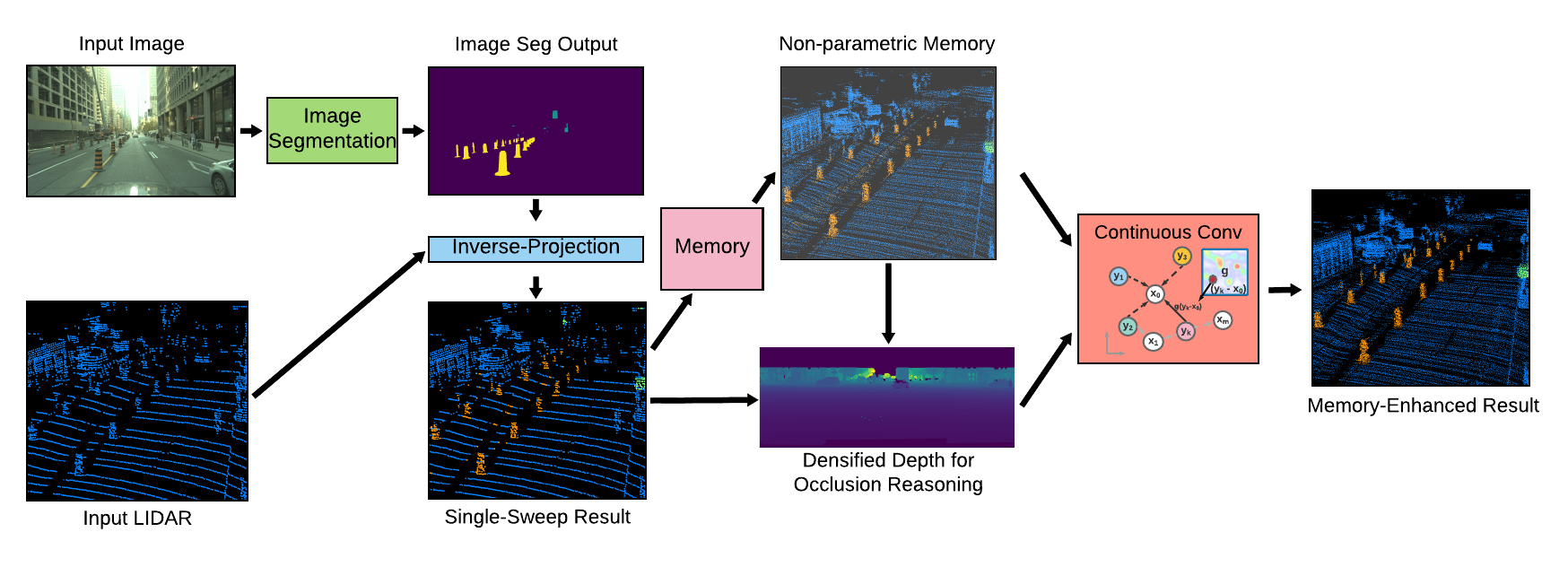}
    \end{center}
    \vspace{-0.6cm}
    \caption{Our method takes as input an image captured from the camera, which is passed through a segmentation model. The resulting pixel-wise labels are re-projected onto the corresponding LiDAR point cloud. We generate a depth map for occlusion reasoning over the past observations in memory. The information in the memory and the current single sweep data are then processed by a continuous convolution model. The memory is subsequently updated with points and image segmentation result from the current sweep.}
    \label{fig:pipeline}
    \vspace{-0.2cm}
\end{figure*}

\section{Learning to Forget/Reinforce/Remember}

In this section, we describe the motivation for and implementation of our non-parametric memory-based model that is able to maintain a record of segmented objects of interest, and make appropriate updates as new data becomes available.

We demonstrate the effectiveness of our approach on the segmentation of construction elements such as cones and barrels, as well as traffic signs. These small objects are critical to safe autonomous driving.
 Note that this method can be easily extended to other classes. 

\subsection{3D Segmentation from Images and LIDAR}

Image segmentation is a well understood area, where numerous works have achieved very impressive performance. 
This is enabled by the dense information captured by the camera. Moreover, the rich texture and colors in images allow models to disambiguate between different surfaces that may otherwise have similar physical structure. 
However, autonomous driving requires knowledge of the surroundings in 3D. LiDAR measurements provide a sparse set of points per sensor sweep that are localized in the 3D world, but provide only sparse information. 
In our work, we propose to leverage the discerning power of an image segmenter while using 3D LiDAR points accumulated over time as a localizer to segment objects with high accuracy in 3D. 

Towards this goal, at each timestep, we first obtain a pixel-wise segmentation on the 2D image captured by a front-facing camera mounted on top of the self-driving car. 
The segmentation model is a deep convolutional neural network using the ResNet-101 \cite{he2016deep} backbone. 
Furthermore, we use a spatial feature pooling scheme based on the PSPNet~\cite{zhao2017pyramid} architecture.
In particular, we follow the standard ResNet-101 backbone's 2048-dimensional output with three large receptive field size pooling operations of $5 \times 5$, $10 \times 10$, and $25 \times 25$ followed by a point-wise convolution for the original output and all three pooling outputs to reduce the output dimension to 512 before concatenation. 
This is processed by an additional 3 ResNet blocks to fuse the multi-scale information. We then use 3 transposed convolution layers with a ResNet block following the first two to upsample the feature volume to input resolution while reducing the feature dimension. Using the camera calibration matrix, we project the 3D points of the corresponding LiDAR sweep onto the camera image. For the 3D points that fall within the image, we associate the image segmentation output at the nearest pixel with the point. 

This naive implementation suffers from a number of shortcomings. First, the LiDAR points collected at different distances within the aggregation time window are considered equally, while segmentation results are less noisy at shorter distances. 
As well, points that were previously labelled as classes of interest keep their labels, even when later sweeps show contradictory information. 
The semantic label given to the earlier point should be removed if at a later time we have a clear line of sight to structures behind the point. This new observation may indicate that the earlier point belongs to a moving object, whose location is now unoccupied at the later time. 
Alternatively, it may be that the earlier foreground label is a false positive. Unfortunately, the method described here is unable to remove these false positives. 
Finally, it is observed that the resulting output in the 3D space is noisy, as slight misalignments between the image segment boundaries and the 3D projection cause large mislabeled regions in the 3D world.

\begin{table*}[t]
\centering
\setlength{\tabcolsep}{3pt}
\begin{tabular}{@{}c|ccc|ccc@{}}
 \multicolumn{1}{c}{} & \multicolumn{3}{c}{Validation IoU} & \multicolumn{3}{c}{Test IoU} \\
\toprule
Model & Traffic Sign & Construction & Mean  & Traffic Sign & Construction & Mean \\
\midrule

Image Baseline & $42.3\%$ & $52.3\%$ & $47.3\%$ & $42.1\%$ & $42.1\%$ & $42.1\%$ \\
Voxel NN~\cite{chris_voxel} & $37.6\%$ & $42.6\%$ & $40.1\%$ & $37.9\%$ & $52.8\%$ & $45.4\%$ \\
Continuous Convolution - Single Sweep & $56.6\%$ & $63.2\%$ & $59.9\%$ & $52.9\%$ & $61.6\%$ & $57.3\%$ \\
\midrule
Continuous Convolution with Memory (Ours) & $59.1\%$ & $66.1\%$ & $62.6\%$ & $56.0\%$ & $64.4\%$ & $60.2\%$ \\
\bottomrule
\end{tabular}
    \vspace{-0.2cm}

\caption{IoU metrics for our model and the comparison.}
\label{table:iou}
    \vspace{-0.5cm}

\end{table*}

\subsection{Non-parametric Memory for Spatio-temporal Segmentation}

We propose a learned model to address the aforementioned shortcomings. The overall pipeline of our model is depicted in (Figure~\ref{fig:pipeline}). 
We use a non-parametric memory structure to maintain our current best estimate for segmentation of aggregated point clouds. The memory stores a length $N$ list of 3D points described by a $\mathbb{R}^3$ coordinate. Additionally, we store a vector $M \in [0, 1]^C$ of $C$ associated class-wise probability estimates for each point indexed by $i$ collected at all time steps up to the current time, containing the classification probability output of the image segmenter. 

Concretely, we define a dynamic graph $G = (P, E)$ where $P$ is the set $\{u\}$ of all historical and current 3D points, and $E$ is the set of edges $\{(u,v) | v \in \text{NN}_K(u)\}$, where the $\text{NN}_K(u)$ is the $K$-nearest spatial neighbors to a point $u$. 
We then use a model based on the continuous convolutions (CC) architecture~\cite{wang2018deep} to process each point. 
While the traditional convolutional neural network operates on discrete locations in a regular grid, the CC method allows input and queried output locations at arbitrary continuous coordinates. 
This bypasses the need to balance between the high memory requirements of a voxelized representation needed by discrete convolutions and the resolution (hence precision) in the output space. 

To appropriately aggregate memory information of estimated classification probabilities, we need to reason about whether or not past observed regions are occluded in the current frame. 

Therefore, we compute occlusion information from the current frame. This requires a dense depth map constructed for the current time step. 

While the density of LiDAR points vary greatly with distance, it is constant in polar coordinates as the pitch (vertical) angular separation of LiDAR beams and the azimuth (horizontal) angular resolution are fixed. 
We use the vehicle pose at each time step to convert both current and previous 3D points to polar coordinates in the current perspective reference frame. 
For each point's coordinates $p = [x, y, z]^T$ where the $x, y, z$ axis point in the forward, right, and up directions from the vehicles, respectively, we compute the point's polar coordinates as

\[
r = \sqrt{x^2+y^2+z^2}, \phi = \text{tan}^{-1} \frac{y}{x}, \theta = \text{sin}^{-1} \frac{z}{r}
\]

where $r$, $\phi$, and $\theta$ are the range, azimuth, and pitch angles, respectively. The remaining gaps are filled in with nearest neighbor interpolation. Finally, we produce an occlusion score for each of the previous 3D points by computing the difference in depth:
\[
o_\text{p} = r_\text{depth image, p} - r_\text{p}
\]
where $r_\text{p}$ is the distance to a previous 3D point queried and $r_\text{depth image, p}$ is the value in the depth image corresponding to the same angular coordinates as the query point. 

Finally, we concatenate the memory contents, the occlusion score, intensity of the LiDAR measurement, and the distance to the vehicle at which each point was measured. This results in a  $D = C + 3$ dimensional feature vector $f$ for each point. 

To aggregate spatial and temporal information for a point $u_i$, we find its $K=50$ nearest neighbor coordinates $\{v_j\}$, and look up their features $\{f_j\}$. This is given as input to a continuous convolution model that consists of four layers with residual connections. For each layer, let us define $N$ as the number of points, $F$ and $O$ as the input and output feature dimensions. The output feature vector's elements are computed as: 

\[
h_{k,i} = \sum_d^F \sum_j^K g_{d,k}(u_i - v_j)f_{d, j}
\]

where $g = \text{MLP}(z; \theta) : \mathbb{R}^3 \rightarrow \mathbb{R}$ is a learned multi-layer perceptron with parameters $\theta$ that transforms a spatial offset $z$ into a scalar weight. 

For our model, the first CC layer expands the feature dimensionality to $16$, while the remaining layers maintain this dimensionality to produce a $C$ dimensional classification output. 
Within each layer, an efficient and equivalent variant of the aforementioned CC operation is used. 
Here, a 2-layer multi-layer perceptron generates kernels based on the relative coordinate offsets between the queried point and its neighbors. 
This is multiplied by the point-wise features transformed by a learned projection, added to the input for the residual connection, and passed to the next layer. 
Additionally, we apply batch normalization at the output of each CC layer. 

The output of this model is the newly updated estimates for the classification probabilities for each 3D point. 

\section{Experimental Evaluation}

In this section, we evaluate our method and compare it with the baseline as well as various ablation studies.

\paragraph{Dataset - ground truth generation} 
We select the construction elements and traffic sign classes as the focus of our technique due to their importance for autonomous driving and relative difficulty for traditional approaches arising from their small physical size. 
Human labelling efforts of point-wise LiDAR point clouds is expensive, as complex and sparse 3D points are difficult to visualize and select. 
Instead, we employ the same 2D semantic segmentation model and aggregation for automatic ground truth generation. 
For this, we train the image semantic segmentation model on a dataset of 10k training, 1.3k validation, and 1.3k testing images with pixel-wise semantic segmentation annotation collected in various North American cities and highways. 
The model is trained on 4 NVIDIA Titan Xp GPUs with ADAM and a learning rate of $1e-5$ until convergence at 240k iterations. 
This method achieves 64.1\% and 63.3\% pixel-wise intersection over union on the traffic sign and construction elements classes in the validation set, respectively. 

Our joint LiDAR and camera image dataset consists of $3141$ training, $430$ validation, and $606$ testing sequences of LiDAR sweeps and associated RGB front camera images. 
Each sequence is defined by a key frame, and contains frames sampled at 0.5 meters of vehicle displacement (or 0.1 seconds if the vehicle moves farther between frames than 0.5 meters). 
To enable training of the memory system, the sequence includes frames prior to the key frame sampled over a displacement of 30 meters. 
The locations of the 3D points are expressed in a globally consistent coordinate system  using an online pose estimation system. 

\begin{figure*}
    \begin{center}
        \includegraphics[width=0.99\linewidth]{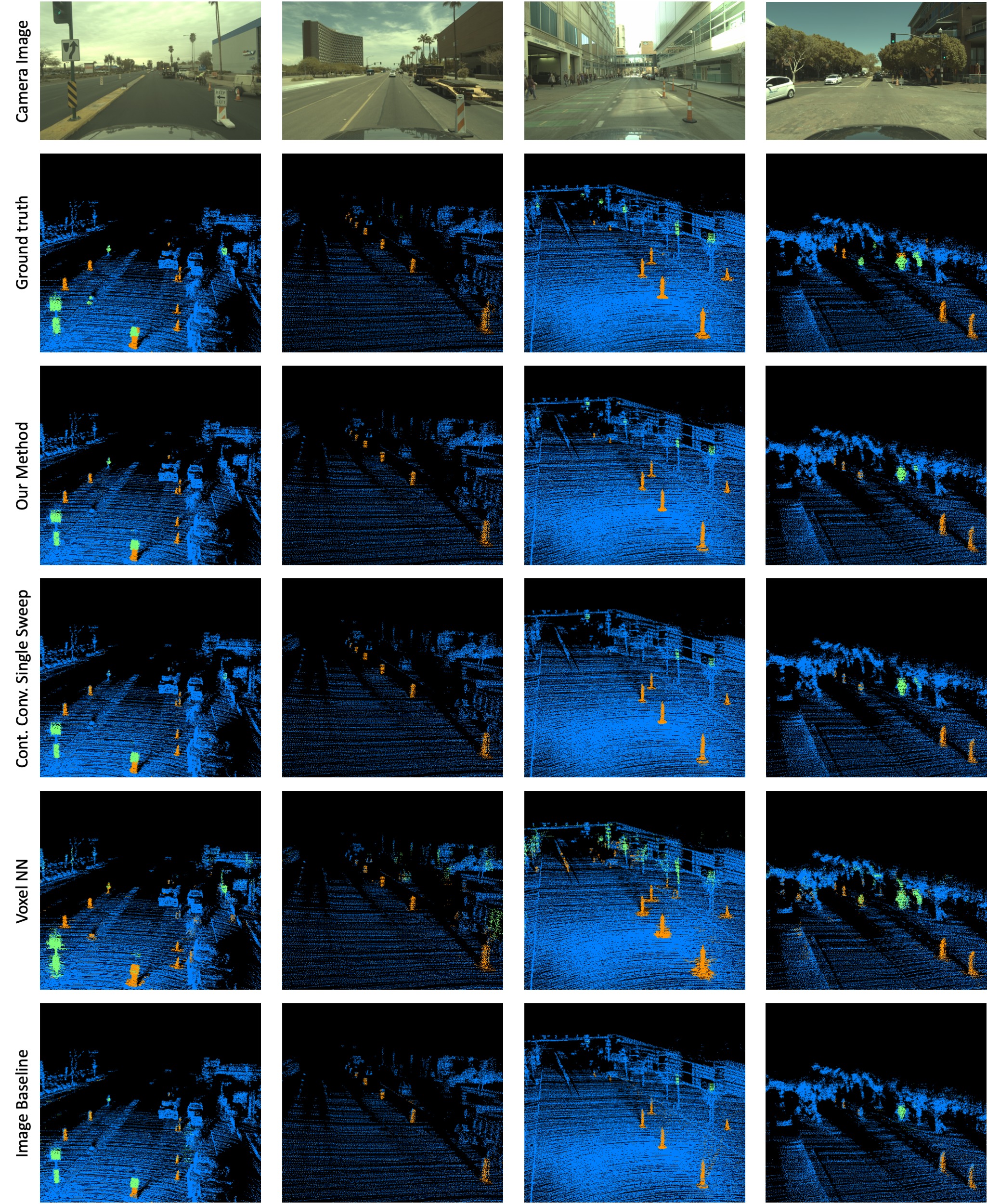}
    \end{center}
   \vspace{-0.3cm}
    \caption{Output comparisons on validation set. In each column, we show an RGB camera input, the automatically generated ground truth, output from our method, output from the continuous convolution model with a single LiDAR sweep as input, output from \cite{chris_voxel}, and output from the image segmentation baseline. Traffic signs are shown in green, while construction elements are shown in orange. This comparison is best viewed at a higher zoom level.}
    \label{fig:full_results}
\end{figure*}

We extend the data sequence with observations for 100 meters with the aforementioned sampling strategy. 
For each frame, the LiDAR points are labeled with the 2D semantic segmentation result. To reduce noise, we select only points within 25 meters of the vehicle. 
By nature, construction cones and signs are largely isolated objects that are easily separated from its environment in 3D. 
Thus, we use the density-based spatial clustering \cite{ester1996density} as a denoising step over the initial point-wise semantic labels.
Following this, we produce a training sample / ground truth pair by using only the sequence of frames over the 30 meters preceding the key frame, while removing the distance filter. 
We use the denoised ground truth results to label this set with k-nearest neighbor search for points that are measured at beyond 25m. We observe in the second row of Fig \ref{fig:full_results} that this method is able to provide us with highly accurate ground truth segmentations.

\begin{figure}
    \begin{center}
        \includegraphics[width=1.0\linewidth]{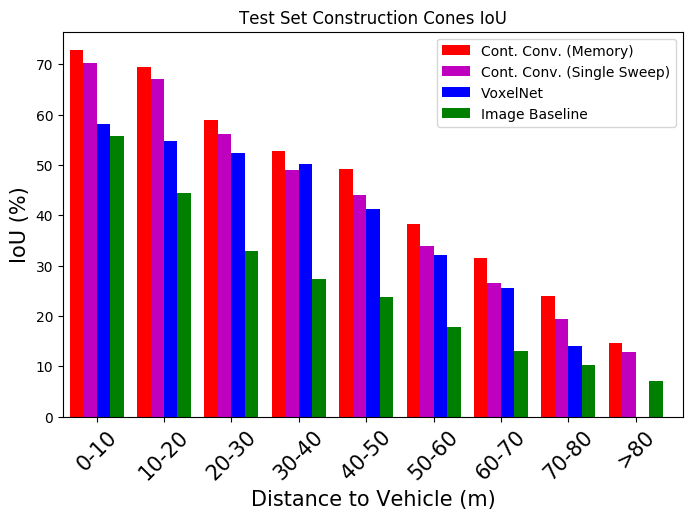}
        \includegraphics[width=1.0\linewidth]{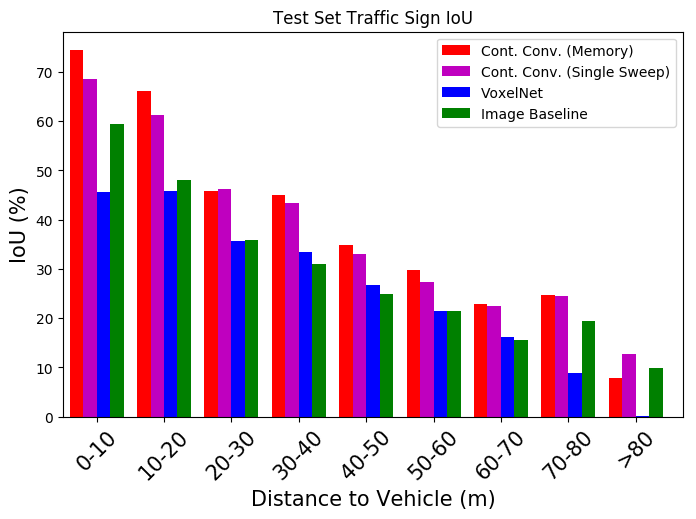}
    \end{center}
  \vspace{-0.4cm}
    \caption{Test set IoU for construction elements (top), and traffic signs (bottom).}
    \label{fig:test_iou}
  \vspace{-0.7cm}
\end{figure}

\paragraph{Training} 
We estimate the sample mean and variances for the $C+3=6$ point-wise features and normalize them relative to each other. This is used to standardize the features, which are then used as input to the continuous convolution model. The model parameters are randomly initialized. At the output, we minimize the average of the standard cross-entropy loss at each point. 

We train our model with a batch size of 1 and the ADAM \cite{adam} optimizer with a learning rate of $1e-3$ on a single NVIDIA Titan Xp GPU until convergence as determined by the validation set at 28k iterations. 
We then fix the batch normalization statistics, and further finetune the model at a learning rate of $1e-4$ for an additional 64k iterations until convergence. 
\paragraph{Metrics} 
We use standard point-wise classification intersection-over-union (IoU) for evaluating our method and the comparison. Moreover, we examine the performance differences between the baselines and our method at different distances from the vehicle. As this model is to be used in the online driving setting, we evaluate over only the LiDAR points that fall within the forward camera view at the final timestep. This has the effect of increasing the difficulty, as the elements in this region have less accumulated measurements than elements next to and behind the vehicle. 
\paragraph{Baselines} 
We analyze the performance of our model with two baselines. The first baseline consists of using the same 2D semantic segmentation model to label the 3D points projected into the image, which we call the \emph{Image Baseline}. Additionally, we implement and train one of the recent fast point cloud segmentation techniques~\cite{chris_voxel} according to the settings described in the paper. We refer to this technique as Voxel NN. 
This model uses 2D convolutions on a simple occupancy grid representation to produce semantic segmentation predictions for 3D point clouds. Specifically, they discretize a given point cloud into a $ W \times H \times Z $ voxel grid, and compute predictions for each voxel using a 2D U-Net model by treating the $ Z $-dimension as the feature channel. 
Per-point predictions are then recovered via nearest neighbour interpolation.
\paragraph{Ablation} 
We also present results for a model where our memory approach is removed from the continuous convolution framework. This model uses a single sweep of the LiDAR point cloud and the corresponding image segmentation. We train this model in a similar fashion as our proposed model, with the exception that a single frame is used as an example instead of an accumulated sequence.  
\paragraph{Quantitative results} 
Table~\ref{table:iou} show the validation and test dataset IoU metrics for traffic signs, construction elements and the mean for all models including the baselines, and the single sweep continuous convolution baseline. 
We observe that our continuous convolution approach outperforms results from other 3D processing methods~\cite{chris_voxel}. 
Moreover, the use of the non-parametric memory presented in our paper consistently improves the results across the validation and test sets both for the traffic signs and the construction elements. 

We further dissect the results by slicing them into object bins according to distance from the vehicle, as shown in Figure~\ref{fig:test_iou} for the test set. 
It is evident that the segmentation performance at larger distance is lower, as the image segmentation is noisier while the LiDAR sweeps return fewer points.
These results also show that our method consistently performs better compared to the previous work, the image baseline, as well as a version of our method without the memory. 
The significant performance gap between our method and the baselines highlights the effectiveness of our model in appropriately aggregating noisy spatio-temporal information especially at range.

\paragraph{Qualitative results} 
Fig \ref{fig:full_results} shows the qualitative comparisons on the validation set across the baseline methods, single LiDAR sweep ablation study, and our proposed method. 
Note that the automatic ground truth generation procedure described above is able to generate accurate ground truth point-wise semantic segmentation for the classes of interest. 
Additionally, we observe that the single LiDAR sweep continuous convolution method without the non-parametric memory tends to miss small elements that are far away, for example the distant traffic sign on the left in first column. 

In comparison with the baseline methods, our model achieves much better performance. The Voxel NN ~\cite{chris_voxel} model's output is far noisier, and often overestimates the extent of the foreground objects. This is likely due to the limited resolution after discretization of the voxel-based model, which is necessary for the model to avoid exceeding memory and processing constraints. 
The noisy object boundaries can have a significant impact on the autonomous vehicle's motion planning systems. In contrast, the continuous convolutions operation avoids the decrease in resolution by directly operating on the 3D points, and produces very accurate foreground extents. 

Finally, the baseline method where the image semantic segmentation output is directly projected onto the LiDAR sweeps often results in trails of foreground labels on background points at occlusion boundaries. 
The construction signs in the first column and the closest cone in the third column show this issue most prominently. Again, the effect on autonomous vehicle operation can be significant, as drivable regions would be labelled as obstacles. In contrast, our method is able to remove these artefacts. 

\paragraph{Performance} 
In our experiments, we used a relatively large and unoptimized image semantic segmentation technique with a runtime of approximately 400ms on a single GPU. The continuous convolution over the non-parametric memory module generally runs in under 20ms, with the exact runtime dependent on the number of foreground points in a scene. Therefore, with an optimized semantic segmentation method that runs at ~130ms, our method will be running at 150ms end-to-end.

\paragraph{Limitations} One area of potential future improvement is as follows. If there are false negatives in the instance segmentation model, the corresponding LiDAR points will not be passed to the continuous convolution model and hence be absent in the foreground estimation. Additionally, there may be elements of interest outside of the LiDAR return range, which would not be registered. Finally, the proposed method's reliance on the image segmenter suggests that its performance may be limited in the dark.


\section{Conclusion}

In this paper, we introduced a non-parametric memory representation for spatio-temporal segmentation. We demonstrated that this representation is effective in aggregating information in local neighborhoods of 3D LiDAR points as well as observations over time to increase robustness and reduce noise in the segmentation output. 

This is made possible by our representation's three important capabilities: (i) it {\it remembers} what it has seen in the past, (ii) it {\it reinforces} or (iii) {\it forgets} its past beliefs based on new evidence. 

In the future, we aim to explore more methods for sensor fusion within a multi-task network that learns joint features for image and point cloud. We also plan to extend our method to other semantic classes of interest. 

\bibliographystyle{unsrt}
\bibliography{egbib}
\end{document}